\PassOptionsToPackage{table}{xcolor}
\documentclass[utf8]{FrontiersinHarvard} % for articles in journals using the Harvard Referencing Style (Author-Date), for Frontiers Reference Styles by Journal: https://zendesk.frontiersin.org/hc/en-us/articles/360017860337-Frontiers-Reference-Styles-by-Journal
\usepackage{url,hyperref,lineno,microtype,subcaption}
\usepackage[onehalfspacing]{setspace}
\usepackage{soul} % for the command \hl
\renewcommand{\hl}[1]{#1}
\usepackage{booktabs}
\usepackage{multirow}
\usepackage{caption}

\def\keyFont{\fontsize{8}{11}\helveticabold }
\def\firstAuthorLast{Jana {et~al.}} %use et al only if is more than 1 author
\def\Authors{Abhishek Jana\,$^{1,*}$, Moeumu Uili\,$^{2,5,*}$, James Atherton\,$^{5}$, Mark O'Brien\,$^{3}$, Joe Wood\,$^{4}$, and Leandra Brickson\,$^{1,5}$}
\def\Address{%
$^{1}$Colossal Biosciences, Dallas, TX, USA\par
$^{2}$Colorado State University, Department of Fish, Wildlife, and Conservation Biology, Fort Collins, Colorado, United States\par
$^{3}$BirdLife International, Cambridge, England\par
$^{4}$Toledo Zoo, Toledo, Ohio\par
$^{5}$Samoa Conservation Society, Samoa.
}

\def\corrAuthor{Moeumu Uili}
\def\corrEmail{moeumu.uili@colostate.edu}

\begin{document}

\onecolumn
\firstpage{1}

\title[An Automated Pipeline for Few-Shot Bird Call Classification]{An Automated Pipeline for Few-Shot Bird Call Classification, a Case Study with the Tooth-Billed Pigeon} 

\author[\firstAuthorLast ]{\Authors} %This field will be automatically populated
% \address{} %This field will be automatically populated
% \correspondance{} %This field will be automatically populated
\address{}
\correspondence{}

% \extraAuth{}% If there are more than 1 corresponding author, comment this line and uncomment the next one.
\extraAuth{Abhishek Jana \\ abhishek@colossal.com}

\maketitle
\begin{abstract}

%%% Leave the Abstract empty if your article does not require one, please see the Summary Table for full details.
\section{}
This paper presents a largely automated one-shot bird call classification pipeline, incorporating targeted manual quality control steps, designed for rare species absent from large publicly available classifiers like BirdNET and Perch. While these models excel at detecting common birds with abundant training data, they lack options for species with only 1--3 known recordings, a critical limitation for conservationists monitoring the last remaining individuals of endangered birds. To address this, we leverage the embedding space of large bird classification networks and develop a classifier using cosine similarity, combined with filtering and denoising preprocessing techniques, to optimize detection with minimal training data. We evaluate various embedding spaces using clustering metrics and validate our approach in both a simulated scenario with Xeno-Canto recordings and a real-world test on the critically endangered tooth-billed pigeon (Didunculus strigirostris), which has no existing classifiers and only three confirmed recordings. The final model achieved 1.0 recall and 0.95 accuracy in detecting tooth-billed pigeon calls, making it practical for use in the field. This open-source system provides a practical tool for conservationists seeking to detect and monitor rare species on the brink of extinction.

\tiny
 \keyFont{ \section{Keywords:} Deep Learning, Bird Conservation, Bird Call Detection, Few-shot Learning, Tooth-billed Pigeon} 
 %All article types: you may provide up to 8 keywords; at least 5 are mandatory.
\end{abstract}
 
\section{Introduction} % --------------------------------------------------
\label{sec:intro}
Effective avian conservation relies on the ability to identify, monitor, and track species populations in their natural habitats. These efforts provide essential data on species distribution, habitat use, and the impact of human activity, informing conservation priorities \citep{Sutherland2004}. For critically endangered species, the need to locate the few remaining individuals is especially urgent. Conservationists may seek to find these individuals for relocation to captive breeding programs or to collect genetic material for sequencing or cryopreservation, ensuring the species' genetic lineage is not permanently lost.

Locating these critically endangered avian species presents significant challenges. Many birds in forest environments reside in the upper canopy, limiting direct visual observation. Camera traps face similar constraints, as their effectiveness is reduced by dense foliage and their fixed field of view. In contrast, passive acoustic monitoring provides a more viable solution, as sound can propagate through vegetation and be recorded omnidirectionally \citep{Marques2013}.

Moreover, when only a few individuals remain, the search area must be extensive, requiring widespread deployment of recording devices. Detecting a single call may necessitate analyzing hundreds or thousands of hours of audio, creating a data volume that is impractical for manual review.

In response to these needs, recent advancements in \hl{artificial intelligence (AI)} have led to the development of bird call classifiers using audio data. On the foundation of earlier methods and datasets \citep{Stowell2016, Vellinga2015}, the BirdCLEF competition emerged to challenge the AI community to accurately classify bird calls \citep{Goeau2014}. This challenge was built on the Xeno-Canto dataset, which continues to grow with more diverse soundscapes and more species each year \citep{Vellinga2015, Kahl2019, Kahl2020, Xenocanto2024}. 

The BirdCLEF challenge has spurred many impressive models in the subsequent years. The most notable models with high accuracy are Perch \citep{Google2023}, which achieved the highest Top-1 accuracy across a benchmarking test \citep{Ghani2023}. Another notable model is \hl{BirdNET 2.4} \citep{Kahl2021}, which was second highest in the same benchmark test \citep{Ghani2023}. These models are trained on large, publicly available datasets, such as Xeno-Canto, and other crowd-sourced bird call sources. However, these datasets often suffer from significant class imbalance, with recordings heavily skewed toward common species \citep{Goeau2018}. This skewed distribution affects model generalization, particularly for species with few or no training examples.

Moreover, while BirdNET has been successfully applied to numerous conservation efforts for species with sufficient training data \citep{Manzano2022, Doell2024, Cole2022, Tilson2022}, it faces challenges in extremely data-scarce scenarios. It struggles with overlapping calls in complex soundscapes, often misclassifying species with acoustically similar vocalizations \citep{Kahl2021}, and also has difficulty detecting soft calls \citep{Priyadarshani2018}. As a result, BirdNET is not always reliable for conservation-driven applications where only a handful of recordings exist and high accuracy is critical \citep{Stowell2019, Perez2023}.

These limitations underscore the need for methods capable of detecting species with minimal training data. Recent advancements have explored zero-shot and few-shot classification to address this challenge, though current methods still fall short of the accuracy required for robust field deployment \citep{Moummad2024, Poutaraud2023}. Some studies have approached zero-shot classification by integrating metadata, such as textual descriptions of bird calls and other relevant features \citep{Gebhard2024, Miao2023}. Notably, the CLAP model has demonstrated promising quantitative performance \citep{Miao2023}. However, it faces challenges in distinguishing fine-grained species-level categories due to its reliance on textual descriptions of the calls \citep{Miao2024}. Alternative approaches have focused on analyzing the embedding space of bird call classifier models \citep{Ghani2023, Sanchez2024}. Among these, one study reported particularly strong results by retraining the final layer using traditional few-shot supervised transfer learning, achieving approximately 0.9 \hl{area under the curve (AUC)} for classification on several bird benchmarks with only four training examples per class \citep{Ghani2023}. A complementary line of work frames rare-call detection as an agile modeling problem, in which embedding-based vector search retrieves acoustically similar segments from a single query example and a human-in-the-loop active learning loop iteratively refines the recognizer \citep{Dumoulin2025}. This interactive retrieval approach has been applied to conservation settings, including a Christmas Island occupancy-modeling case study that is biologically analogous to our own. Our method differs in that it defines a fixed cosine similarity threshold from the available recordings, without an interactive retrieval or relabeling loop, trading the adaptivity of active learning for a simpler, fully specified pipeline that can be deployed once calibrated.

Inspired by this work, our study aims to refine one-shot or few-shot classification accuracy for extremely rare bird species, particularly when the available field data consist of only a single 5-minute recording of an individual. The proposed pipeline is developed to classify a specific bird of interest for targeted surveys. To achieve this, the data is carefully preprocessed to extract the specific features relevant to the particular bird of interest. Then, a deep learning model is selected using rigorous clustering-quality metrics to select a state-of-the-art bird call classification model to use for embeddings. Finally, a cluster thresholding method that minimizes false negatives is developed to provide the final classification from the embeddings. The largely automated pipeline for this technique, with targeted manual quality control steps, is made publicly available in a repository to support conservation efforts in tracking and detecting rare species.

This method is first validated using a simulated dataset of bird species from the American Northeast, providing a sufficiently large test set to strengthen the reliability of the results. We then apply the approach to a real-world scenario: detecting the tooth-billed pigeon (\textit{Didunculus strigirostris}), a critically endangered species native to Samoa whose continued existence in the wild is uncertain. Only three confirmed field recordings of this species are currently available, alongside numerous recordings of other bird species sharing its habitat. Evaluating our system in this challenging context allows us to rigorously demonstrate its utility for conservation efforts targeting extremely rare birds.

The contributions of this work are as follows:

\begin{itemize}
    \item \textbf{Targeted Detection Pipeline:} A specialized pipeline focused on maximizing the detection of a single bird species, with the intent to improve recall significantly over networks trained to detect multiple bird species. 
    
    \item \textbf{Refinement of Bird-Call Embeddings:} While prior work by Ghani et al. \citep{Ghani2023} has explored bird-call embeddings, our study revisits bird call classification models, assessing them with three clustering metrics to identify the approach that minimizes intra-class variability, a factor known to improve the performance of few-shot transfer learning \citep{Chen2019}.
    
    \item \textbf{Real-World Application:} Our results demonstrate the detection pipeline's effectiveness, tested on real-world field recordings from the Samoan Conservation Society.
    
    \item \textbf{Public Repository:} We provide a largely automated, end-to-end pipeline with targeted manual quality control steps, fully accessible through a public repository, allowing conservationists and researchers to easily train and deploy new classifiers for their bird species of interest. \url{https://github.com/abhishek-jana/few-shot-bird-call} The repository is archived on Zenodo (\url{https://doi.org/10.5281/zenodo.20726966}) under the MIT license.
    
\end{itemize}

%\vspace{-4mm} 
%\vspace{-12pt}
\section{Methods} % ---------------------------------------------------
\label{sec:methods} 

\begin{figure*}[b]
    \centering
    \includegraphics[width=\linewidth]{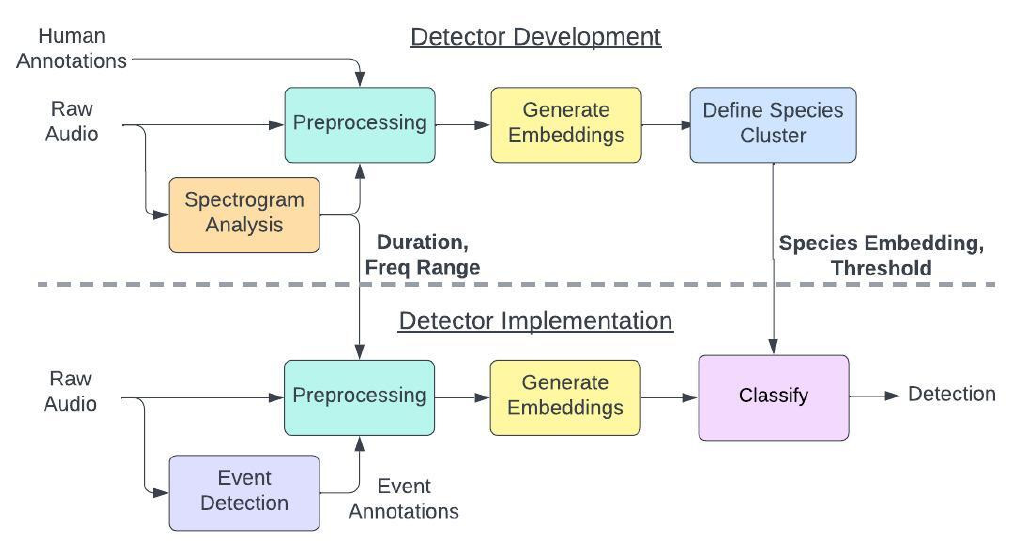}
    \caption{Overview of the classification pipeline. The upper section illustrates the development process, where key parameters such as duration, frequency range, and threshold are defined, along with the species embedding for the target species. The lower section demonstrates how the finalized model is applied on field data to detect calls from the species of interest.}
    \label{fig:architecture}
\end{figure*}

Our methodology first preprocesses raw audio recordings, followed by feature extraction using a deep learning model to generate embeddings. These embeddings are then compared to a species-specific reference embedding to classify calls based on similarity. Using this approach, the model is fine-tuned to detect the target species amidst acoustically similar bird calls. To utilize this model, an event detector first isolates bird calls from the entire recording, after which classification is performed using the developed pipeline. Figure \ref{fig:architecture} provides an overview of the entire pipeline development and implementation. The following sections describe each component of this workflow in detail.

\subsection{Eastern Towhee Dataset}
\label{sec:eastern_towhee}
\hl{To evaluate the system, we designed a test scenario that replicates a similar conservation challenge while enabling validation with a larger test set. This scenario was simulated using five common bird species that frequently co-occur in the Appalachian region of the United States: Kentucky Warbler, White-Throated Sparrow, Carolina Wren, Northern Cardinal, and Eastern Towhee. The Eastern Towhee was designated as the rare bird of interest for this study.

For each species, three recordings from Xeno-Canto were used to create the training dataset, while six recordings per species were reserved for testing. Because of the larger test dataset, this dataset provides a broader validation of the classifier. However, unlike the Tooth-Billed Pigeon dataset, it does not evaluate the model’s performance on species absent from Perch or BirdNET’s training data, nor do the recordings share uniform environmental conditions or recording setups.

Through event detection, one recording produces multiple calls for evaluation. However, calls from the same recording likely originate from a single individual and are therefore acoustically similar. To prevent bias, train/test splits were made at the recording level rather than at the individual call level. The number of recordings and detected calls for the Eastern Towhee dataset is detailed in Table \ref{tab:et_dataset}.}

\begin{table}[htbp]
    \centering
    \captionsetup{justification=centering}
    \begin{tabular}{lcccc}
        \toprule
        \multirow[b]{2}{*}{\textbf{Class}} & \multicolumn{2}{c}{\textbf{Train}} & \multicolumn{2}{c}{\textbf{Test}} \\
        \cmidrule(lr){2-3} \cmidrule(lr){4-5}
                       & Calls & Recordings & Calls & Recordings \\
        \midrule
        Kentucky Warbler         & 10 & 3 & 45 & 6 \\
        White-Throated Sparrow   & 17 & 3 & 43 & 6 \\
        Carolina Wren            & 17 & 3 & 64 & 6 \\
        Northern Cardinal        & 10 & 3 & 43 & 6 \\
        Eastern Towhee           & 18 & 3 & 47 & 6 \\
        \bottomrule
    \end{tabular}
    \caption{Details of the Eastern Towhee dataset}
    \label{tab:et_dataset}
\end{table}

\subsection{Tooth-Billed Pigeon Dataset}
\label{sec:tooth_billed}
\hl{To collect the Tooth-Billed pigeon dataset, two types of acoustic recording units (ARUs) were used: Audiomoth and AR4 recorders, both equipped with built-in omnidirectional microphones. In total, fifteen Audiomoth and five AR4 recorders were deployed across randomly selected sites in Samoa in either intact native forest or disturbed open forest. To ensure spatial diversity, ARUs were placed at least 5 km apart and left in the field for three to four months per monitoring period, accumulating three years of recordings from 2021 to 2023. The units collected data year-round, capturing both the wet season (November–April) and the dry season (May–October). Regular maintenance allowed for battery replacements and memory card changes.}

\hl{From the over 20,000 hours of recordings collected, a subset of 160 hours was selected for manual analysis based on data quality. Expert annotators manually reviewed these recordings, identifying and extracting clips containing tooth-billed pigeon calls based on auditory recognition. Across the entire dataset, a small number of recordings containing confirmed tooth-billed pigeon calls were identified. The 4-minute, 49-second recording was reserved for training and contained 36 tooth-billed pigeon calls from a single individual. The test set consists of a separate group of five field recordings provided by our collaborators, described below. All training and test recordings are temporally and spatially distinct, with no temporal or recording-level overlap between training and test data.

These 160 hours of recordings also contained vocalizations from other bird species, most commonly the crimson-crowned fruit dove, white-throated pigeon, Polynesian wattled honeyeater, Samoan starling, and Pacific imperial pigeon. Since these species were frequently detected by our ARUs, they were selected as comparison classes for the classifier. The Pacific imperial pigeon was of particular interest due to its highly similar call structure to the tooth-billed pigeon, as illustrated in Figure} \ref{fig:avg_spectrogram}. \hl{To provide strong counter-examples for the classifier, we selected an additional 20 minutes of recordings containing Pacific imperial pigeon calls from the full dataset. Manual annotations from local experts accompany the full 30 minutes of recordings, marking species identification and the start and end times of bird calls within each segment.}

To define the classification threshold in the embedding space, a training set was constructed using a 4-minute, 49-second field recording of a tooth-billed pigeon, which contained 36 calls from a single individual. Calls from other species present in the same recording were included as comparison classes.

For the test set, five field recordings provided by our collaborators were used for independent evaluation, with durations of approximately 3 minutes, 1 minute 45 seconds, 42 seconds, 10 seconds, and 11 seconds. These recordings contain tooth-billed pigeon calls together with calls from the co-occurring comparison species. From them, 32 candidate call segments were curated and given to expert annotators for confirmation; these 32 segments are a subset of the larger test set detailed in Table~\ref{tab:tbp_dataset}. Two expert annotators independently marked individual calls in these segments, and a segment was accepted as a given species only when both annotators agreed.

Of the five recordings, two contained confirmed tooth-billed pigeon calls, totaling seven calls from up to two individuals (these two recordings are 10 and 11 seconds long, 21 seconds in total), while the remaining three were dominated by acoustically similar co-occurring species, principally the Pacific imperial pigeon. The Tooth-Billed pigeon dataset is detailed in Table \ref{tab:tbp_dataset}. 

\begin{table}[htbp]
    \centering
    \begin{tabular}{lcccc}
        \toprule
        \multirow[b]{2}{*}{\textbf{Class}} & \multicolumn{2}{c}{\textbf{Train}} & \multicolumn{2}{c}{\textbf{Test}} \\
        \cmidrule(lr){2-3} \cmidrule(lr){4-5}
                       & Calls & Recordings & Calls & Recordings \\
        \midrule
    Tooth Billed Pigeon & 36 & 1 & 7 & 2 \\
    Pacific Imperial Pigeon & 27 & 3 & 20 & 3 \\
    White Throated Pigeon & 16 & 3 & 11 & 3 \\
    Crimson Crowned Fruit Dove & 23 & 3 & 20 & 3 \\
    Polynesian Wattled Honeyeater & 22 & 3 & 19 & 3 \\
    Samoan Starling & 5 & 4 & 6 & 3 \\
        \bottomrule
    \end{tabular}
    \caption{Details of the Tooth-Billed Pigeon dataset}
    \label{tab:tbp_dataset}
\end{table}

\subsection{Preprocessing}
\label{sec:preprocessing}

The audio recordings were resampled to 22.05 kHz, enabling analysis of frequencies up to 11 kHz, which covers the typical range of bird calls \citep{dooling2000hearing}.

Recordings were filtered based on the species of interest’s specific frequency range, determined from existing literature or spectrogram analysis, as detailed in later sections. This bandpass filtering was applied consistently to all clips, both training and test, for both the Eastern Towhee and Tooth-Billed Pigeon datasets. The frequency limits differ between species, reflecting each species’ vocal range; the per-species bandpass limits used in this study are listed in Table~\ref{tab:bandpass}. For the Tooth-Billed Pigeon, the spectrogram analysis tool identified a vocal range of 155--674~Hz; a slightly wider filter of 100--700~Hz was applied in practice to avoid clipping call boundaries.

\begin{figure*}[t]
    \centering
    \includegraphics[width=\linewidth]{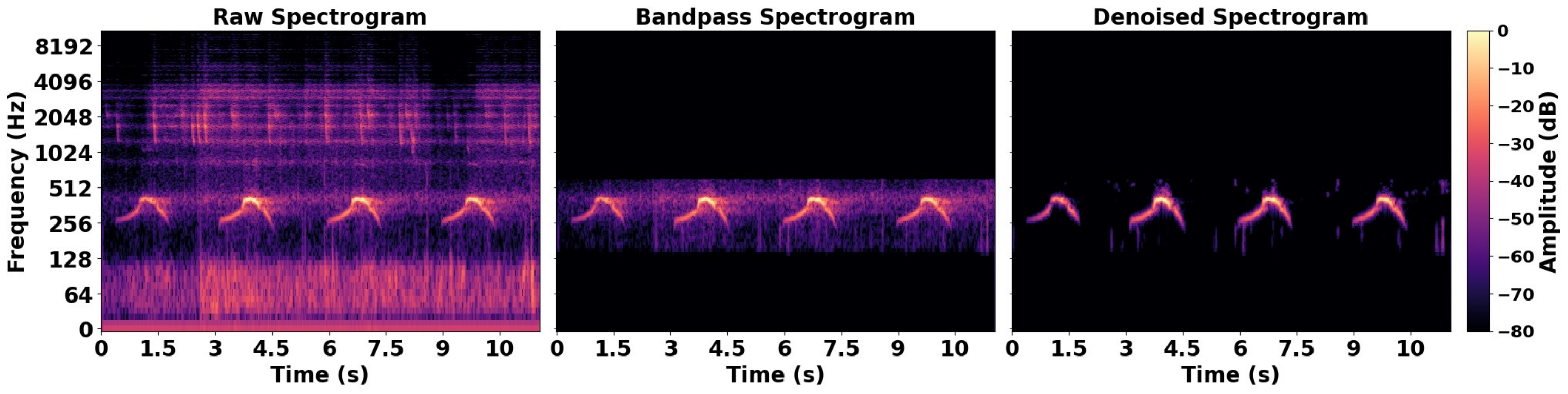}
    \caption{Comparison of raw (left), bandpass filtered (center), and denoised (right) spectrograms of a TBP bird call, shown on a (0-80) dB scale. The bandpassed spectrogram isolates the target frequency band, while the denoised spectrogram highlights essential call features.}
    \label{fig:spectrogram}
\end{figure*}

Next, the noisereduce library \citep{tim_sainburg_2019_3243139, sainburg2020finding} was used to enhance the clarity of the bird calls by reducing background noise. Accounting for background noise has been shown to improve classification performance, particularly in data-sparse scenarios where clean training samples are limited \citep{Lauha2022}. Figure \ref{fig:spectrogram} illustrates a sample Tooth-billed Pigeon spectrogram as it progresses through these initial steps of the preprocessing pipeline.

Each labeled call was then segmented from the full field recording. An automated tool for determining the optimal clip length is discussed in the following subsection. For this study, a 2-second clip length was selected. Calls shorter than this duration were centered and zero-padded. These segmented calls were normalized to -20 dB amplitude.

For the deep learning models investigated in this work, spectrograms serve as the primary input feature. Mel-spectrograms, which are extensively utilized in human speech processing and event recognition tasks \citep{kiyokawa2019sound, delphin2019mean}, were employed for this study. The Mel band count was configured to 227, aligning with the parameter settings used in comparable studies. To ensure compatibility with the specified dimensions (227x227 pixels), the hop length was dynamically adjusted. The Mel-spectrograms were converted to a decibel scale, with values below -40 dB set to zero to filter out low-energy noise and irrelevant sounds.

Finally, manual inspections were conducted on the resulting spectrograms to identify and eliminate any segments containing overlapping or simultaneous vocalizations from multiple species. This ensures that each call accurately represents a distinct vocalization event, thereby allowing for clear delineation of the embedding for each species. In practice, the bandpass filtering and denoising steps applied earlier in the pipeline substantially suppressed out-of-band noise and non-target energy, such that no clips required exclusion due to overlapping vocalizations in either the Eastern Towhee or Tooth-Billed Pigeon datasets used in this study. However, this step remains a potential bottleneck for large-scale deployment in more acoustically complex environments, and the potential for automating it via source separation techniques is discussed in the Discussion.

\subsubsection{Spectrogram Analysis Tool}

A spectrogram analysis tool was created to determine the average call duration and frequency range of bird calls, necessary hyperparameters for the preprocessing steps. This tool can be skipped if these parameters are already known.

\begin{figure}
    \centering
    \includegraphics[width=0.65\linewidth]{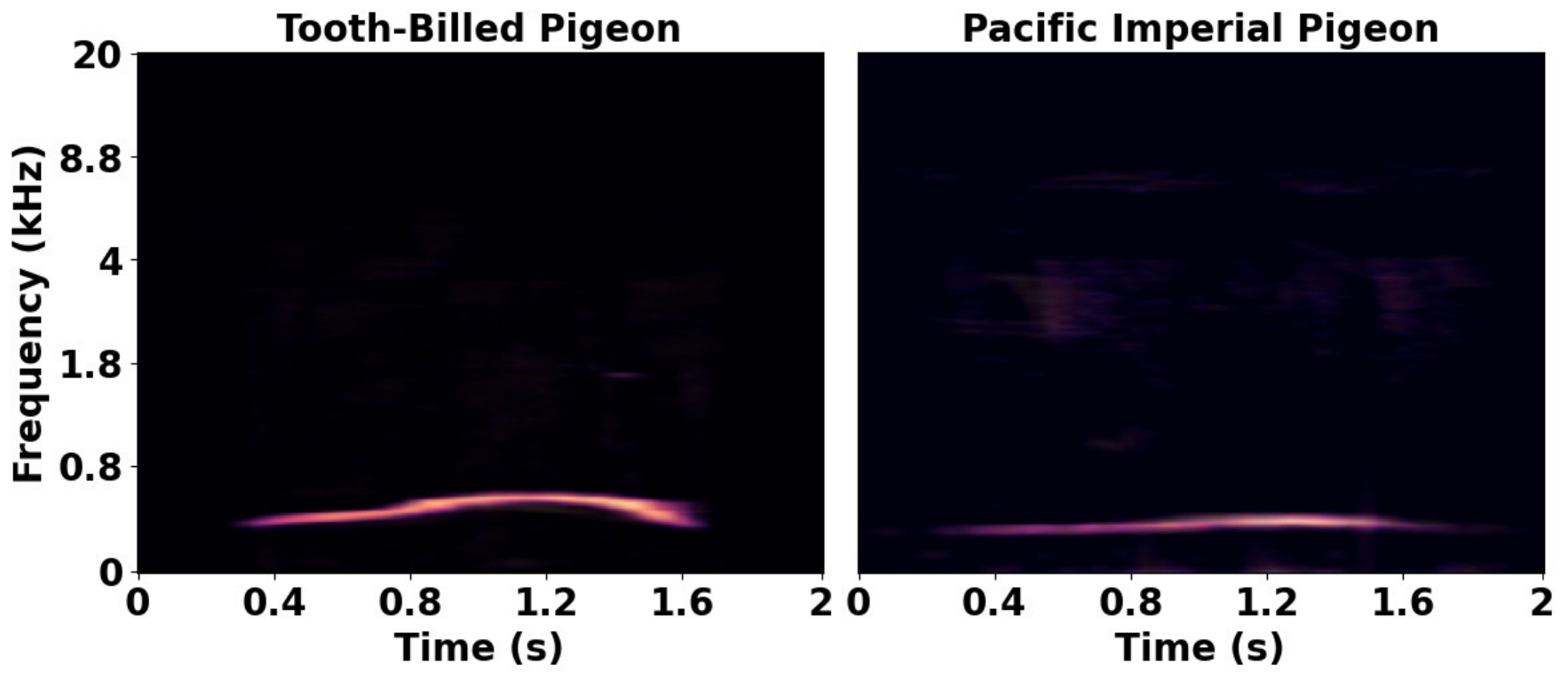}
    \caption{Average spectrograms of vocalizations from the tooth-billed pigeon (left) and pacific imperial pigeon (right), the species in our dataset with a call most similar to the tooth-billed pigeon.}
    \label{fig:avg_spectrogram}
\end{figure}

First, average spectrograms are generated for the labeled calls, as shown in Figure \ref{fig:avg_spectrogram} for the tooth-billed pigeon (left) and the pacific imperial pigeon (right). The minimum and maximum frequencies are identified using basic image processing techniques and then expanded by 50\% to ensure full capture of the call. For the tooth-billed pigeon spectrogram, this resulted in a range of 155 to 674 Hz, which aligns well with documented frequency ranges \citep{beichle1991status, baumann2020acoustical, serra2020using}. 

In a similar way, the average call duration is calculated from the spectrograms, with a 50\% buffer applied to each end. For the tooth-billed pigeon, this results in 1.5 seconds, while the longest call in our study, the Pacific imperial pigeon, spans 1.6 seconds. To ensure that all species' calls are fully included in the clip length, a duration of 2 seconds was chosen for our model.

The bandpass frequency limits applied to each species in both datasets are summarised in Table~\ref{tab:bandpass}. These limits were derived using the spectrogram analysis tool above, with a conservative buffer to avoid clipping call boundaries.

\begin{table}[htbp]
    \centering
    
    \begin{tabular}{llcc}
        \toprule
        \textbf{Dataset} & \textbf{Species} & \textbf{Low (Hz)} & \textbf{High (Hz)} \\
        \midrule
        ET  & Eastern Towhee           & 2500 & 6500 \\
        ET  & White-throated Sparrow   & 2500 & 5000 \\
        ET  & Northern Cardinal        & 1100 & 5000 \\
        ET  & Carolina Wren            & 1400 & 5000 \\
        ET  & Kentucky Warbler         & 2000 & 5500 \\
        \midrule
        TBP & Tooth-billed Pigeon      & 100  & 700  \\
        TBP & Pacific Imperial Pigeon  & 100  & 700  \\
        TBP & White-throated Pigeon    & 100  & 700  \\
        TBP & Crimson-crowned Fruit Dove & 200 & 900 \\
        TBP & Polynesian Wattled Honeyeater & 900 & 4000 \\
        TBP & Samoan Starling          & 1300 & 4200 \\
        \bottomrule
    \end{tabular}
    
    \caption{Per-species bandpass filter limits applied during preprocessing for both the Eastern Towhee (ET) and Tooth-Billed Pigeon (TBP) datasets.}
    \label{tab:bandpass}
\end{table}

\subsection{Generate Embeddings}
Embeddings were generated for all bird calls across a variety of deep learning models to evaluate how well different models distinguish between species. For this section, the goal is to evaluate each model to find one that best clusters the labeled species' calls and reduces intra-class variability. We evaluated 15 models, including 13 convolutional neural networks (CNNs) and 2 transformer-based models, extending previous benchmarking efforts \citep{Ghani2023} by incorporating more recent architectures and evaluation metrics.

CNNs are adept at capturing spatial hierarchies \citep{Lecun2015}, while transformers excel in modeling temporal dependencies \citep{Vaswani2017}, critical for audio analysis. For bird call classification, we utilized models pre-trained on datasets such as ImageNet \citep{Deng2009} and Xeno-Canto \citep{Kahl2021}, including those designed for both spectrogram images and raw audio. The general foundation models selected included ResNet(18,34,50) \citep{He2016}, ResNeXt \citep{Xie2017}, EfficientNet(B0,B1,B2,B3) \citep{Tan2019}, VGG \citep{Simonyan2014}, MobileNet \citep{Howard2017}, and DenseNet \citep{Huang2017}. For transformer-based models, we selected the Vision Transformer (ViT) \citep{Dosovitskiy2020} and Computer Vision Transformer (CVIT) \citep{Roka2023}.

Additionally, we included BirdNET \citep{Kahl2021} and Perch \citep{Google2023}, which are specifically designed for bird call classification and are based on ResNet and EfficientNet-B1 architectures, respectively. These two models have been shown to produce the best embeddings for bird call classification in previous benchmarking tests \citep{Ghani2023}.

All audio clips were converted to 227$\times$227 pixel mel-spectrograms using a dynamically adjusted hop length, as described in Section~\ref{sec:preprocessing}. For the CNN-based models, this single-channel spectrogram was replicated across three channels to match the input expectations of the ImageNet-pretrained networks, while the transformer-based models (ViT and CVIT) used the same 227$\times$227 spectrogram without modification. BirdNET and Perch apply their own internal preprocessing, so for these two models the raw 22.05~kHz audio clips were provided as input and embeddings were extracted from their penultimate layers.

\subsection{Clustering Analysis to Select Model}
\label{sec:clustering}

For both datasets, embeddings used for model selection were generated from the training recordings only, using the same preprocessing pipeline applied to training data: resampling, bandpass filtering, and denoising. For the TBP dataset, this corresponds to the 4-minute, 49-second training recording containing 36 tooth-billed pigeon calls and calls from co-occurring species. For the ET dataset, the Xeno-Canto training recordings for all five species were used.

To make explicit which operations are applied at each stage of the workflow, Table~\ref{tab:stages} summarizes the preprocessing steps used during model selection, threshold definition, testing, and field deployment. The signal-level steps (resampling, bandpass filtering, denoising, and amplitude normalization) are applied identically at every stage, ensuring that embeddings are always generated from uniformly preprocessed 2-second clips. The wavelet event detector differs by stage: the controlled evaluation and the development stages operate on expert-annotated, manually segmented calls, so event detection is used only for training-data curation and during deployment, where call timings are not known in advance.

\begin{table}[htbp]
    \centering
    
    \begin{tabular}{lcccc}
        \toprule
        \textbf{Preprocessing step} & \textbf{Model selection} & \textbf{Threshold definition} & \textbf{Testing} & \textbf{Deployment} \\
        \midrule
        Resampling (22.05 kHz)        & Yes & Yes & Yes & Yes \\
        Bandpass filtering            & Yes & Yes & Yes & Yes \\
        Denoising (noisereduce)       & Yes & Yes & Yes & Yes \\
        Amplitude normalization (-20 dB) & Yes & Yes & Yes & Yes \\
        Wavelet event detection       & No$^{*}$ & No$^{*}$ & No$^{*}$ & Yes \\
        \bottomrule
    \end{tabular}
    
    \caption{Preprocessing steps applied at each stage of the pipeline. $^{*}$Model selection, threshold definition, and testing operate on expert-annotated, manually segmented call clips; the event detector is used for training-data curation and field deployment, where call boundaries are not pre-annotated.}
    \label{tab:stages}
\end{table}

The embeddings generated by the various deep learning models were reduced using principal component analysis (PCA) to increase the information density of the embeddings. For each model, the number of principal components was selected to capture 95\% of the variance in the embeddings. Following the dimensionality reduction, L2 normalization was applied.

After generating the PCA embeddings for each call, we evaluated the clustering quality of the neural network embeddings to select the most suitable model for our detector. We used three metrics to assess clustering effectiveness in the embedding space: the Silhouette score, which measures cluster cohesion and separation \citep{dudek2020silhouette, Sanchez2024}; the Davies-Bouldin index, which evaluates cluster compactness and separation \citep{petrovic2006comparison, merchan2019detection}; and the Dunn index, which considers the ratio of the minimum inter-cluster distance to the maximum intra-cluster distance, rewarding well-separated and compact clusters \citep{dunn1974well}.

To provide a comparative overview, the scores were normalized such that higher values indicate better clustering quality across all metrics. Specifically, the Silhouette and Dunn scores were normalized to a 0-1 scale, and the Davies-Bouldin scores were inverted for consistency. The overall score for each model was then calculated as the average of these normalized scores. This normalized score was used to select the final model for our work. 

%\subsection{Cosine Similarity Based classification}
\subsection{Defining the Species Cluster}
\label{sec:cosine_similarity}

Once a model is selected, the next step is to define a classifier based on the embeddings generated by the model. For this work, we use a cosine similarity-based approach, a common method for comparing embeddings in acoustic classification tasks \citep{Liu2023, Abdullah2022}. \hl{We do not fine-tune or retrain any neural networks in this study. Instead, we leverage the precomputed embedding space from the model selected in the previous section and define a classification threshold based on cosine similarity.}

To construct the classifier, we first calculate the median PCA embedding for each species, representing the centroid of its embedding space cluster. We then compute the cosine similarity between the embeddings of labeled calls and these median embeddings. The resulting similarity scores are transformed using a softmax function to provide a probabilistic interpretation. 

\hl{When dealing with extremely rare species, minimizing false negatives is critical, as even a single missed call can impact conservation efforts. Additionally, because the calls of rare species occur far less frequently than those of other birds, standard metrics such as accuracy become less meaningful. Therefore, we selected recall as the metric for threshold selection} \citep{Sokolova2009}, \hl{as it directly quantifies the model’s ability to correctly identify calls from the species of interest. The final classification threshold is set to the lowest similarity score that achieves a specified recall for the target species in the training set. While this parameter is adjustable, a recall of 0.9 was chosen in this study to account for outliers and potential mislabeling.} Setting the threshold to achieve a fixed recall necessarily accepts some false positives: a lower threshold captures more true calls but also admits more non-target detections. Practitioners should calibrate this tradeoff based on their deployment context, raising the threshold if manual review burden is a concern or lowering it if missing calls is the greater risk.

\subsection{Evaluating the Classifier}
\hl{To assess the performance of the classification model, we evaluated precision, recall, F1 score, and accuracy} \citep{Sokolova2009} \hl{using the set-aside test dataset of tooth-billed pigeon calls. Accuracy was calculated by treating the classifier as binary, measuring the proportion of correctly identified calls from the species of interest relative to all classification decisions.

Additionally, we report the F1 score, a metric that balances precision and recall. The F1 score is especially useful in scenarios with extreme class imbalance, ensuring that both false positive and false negative errors are considered in the evaluation.}

\subsection{Event Detection}
\label{sec:event_detection}
Following model development, rapid labeling of bird call events is essential for accurate segmentation during preprocessing. In field deployment, new data will not have pre-annotated labels or precise call timings, and therefore, an event detector will be necessary. The authors could not find an easy-to-use, publicly available tool suited for bird call event detection, leading us to create and open-source our own. A wavelet-based event detector was developed to use on the band-passed, denoised field data. This detector was inspired by existing wavelet transformation methods for event detection \citep{Juodakis2022, Du2006}. Crucially, applying bandpass filtering and denoising prior to event detection substantially reduces the number of candidate segments passed to the classifier: out-of-band sounds and broadband noise are suppressed before detection, so the detector responds primarily to calls within the target species' frequency range. This pre-filtering step therefore acts as a first-pass filter that eliminates a large proportion of non-target vocalizations before any embedding-based classification is performed.

This event detector starts by identifying the minimum and maximum significant frequencies (typically 20 Hz to 8,000 Hz) using the short-time Fourier transform (STFT). These values guide the wavelet transform scales. Features are computed by summing the absolute values of wavelet coefficients across 20 logarithmically spaced scales, using the Morlet wavelet. A moving average filter with a 0.15-second window smooths the features to enhance continuous event detection.

Detected events are defined as those where smoothed features exceed 10\% of the mean value. Events shorter than 0.5 seconds, the typical call length in our dataset, are filtered out, though this duration can be adjusted for species with shorter calls.

It is important to note that the classification metrics reported in Section~\ref{sec:results} are computed on manually segmented, expert-annotated call clips and therefore do not capture any false negatives introduced at the event detection stage. In field deployment, system-level recall is bounded by the recall of the event detector. The event detector is applied during training data curation and during field deployment (the lower section of Figure~\ref{fig:architecture}), but not during the controlled classification evaluation reported in Section~\ref{sec:results}. We evaluated the event detector against the expert annotations on these recordings (Section~\ref{sec:discussions}, Table~\ref{tab:detector_ablation}): with bandpass filtering and denoising it recovered all expert-annotated calls, including all seven tooth-billed pigeon calls. Because the annotated call set was itself curated with assistance from the detector, a fully independent measurement of detector recall on exhaustively hand-labeled audio, and thereby the true end-to-end system recall, remains a valuable direction for future work.

\section{Results} % -----------------------------------------------------
\label{sec:results}

\subsection{Spectrogram Analysis and Preprocessing}

Figure \ref{fig:avg_spectrogram} displays the average spectrograms for the tooth-billed pigeon and the Pacific imperial pigeon. The automated analysis identified a frequency range of 155 to 674 Hz, closely aligning with documented values (150–650 Hz) \citep{beichle1991status, baumann2020acoustical, serra2020using}. 

\hl{When used on these datasets, this tool effectively determined the frequency ranges for less-documented species and is valuable when the frequency range is unknown. However, while many species yielded clear, coherent average spectrograms, not all provided as distinct a signal as those shown in Figure \ref{fig:avg_spectrogram}.} This discrepancy likely arises from variability in call uniformity or diversity in call types. A similar effect was observed in call duration estimation, though this parameter is less sensitive. To ensure consistency, a call length of 2 seconds was selected.

\subsection{Model Selection for Embeddings}

To determine the most suitable deep learning model for generating call embeddings, we assessed clustering performance using the metrics summarized in Table \ref{tab:cluster_performance}. Perch emerged as the top performer with an overall score of 0.99, followed by BirdNET at 0.72, which also exhibited strong intra-cluster compactness and inter-cluster separation. Based on these results, Perch was selected for further analysis and classification pipeline development. However, if computational efficiency is a priority, BirdNET remains a viable alternative.

\begin{table*}[ht]
    \centering
    \begin{tabular}{lccccc}
    \toprule
    \textbf{Model} & \textbf{Overall Score} & \textbf{Silhouette Score} & \textbf{DB Index} & \textbf{Dunn Index} & \textbf{N Components} \\
    \midrule
    Perch            & \cellcolor{green!25}0.99 & \cellcolor{green!25}0.21 & 1.87 & \cellcolor{green!25}0.41 & 142 \\
    BirdNET          & 0.72 & 0.19 & 2.04 & 0.32 & 130 \\
    ResNet18         & 0.59 & 0.18 & 1.81 & 0.23 & 62 \\
    EfficientNet-B0  & 0.58 & 0.18 & 2.02 & 0.27 & 82 \\
    ResNet50         & 0.54 & 0.16 & 2.05 & 0.27 & 92 \\
    EfficientNet-B2  & 0.53 & 0.17 & 2.11 & 0.26 & 95 \\
    DenseNet         & 0.53 & 0.14 & 2.16 & 0.31 & 76 \\
    EfficientNet-B1  & 0.52 & 0.17 & 1.90 & 0.23 & 76 \\
    VGG              & 0.51 & 0.17 & 1.95 & 0.23 & 87 \\
    ResNet34         & 0.44 & 0.15 & 2.08 & 0.24 & 69 \\
    MobileNet        & 0.43 & 0.15 & 2.19 & 0.26 & 98 \\
    EfficientNet-B3  & 0.42 & 0.15 & 2.20 & 0.25 & 102 \\
    ResNeXt          & 0.37 & 0.13 & 2.17 & 0.26 & 116 \\
    ViT              & 0.33 & 0.11 & \cellcolor{green!25}3.43 & 0.40 & 236 \\
    CVIT             & 0.19 & 0.11 & 3.12 & 0.30 & 141 \\
   % \bottomrule
    \end{tabular}
    \caption{All clustering performance metrics of the various models tested. N Components is the number of principal components used for the reduced embedding.}
    \label{tab:cluster_performance}
\end{table*}

\begin{figure}[t]
    \centering
    \includegraphics[width=0.7\textwidth]{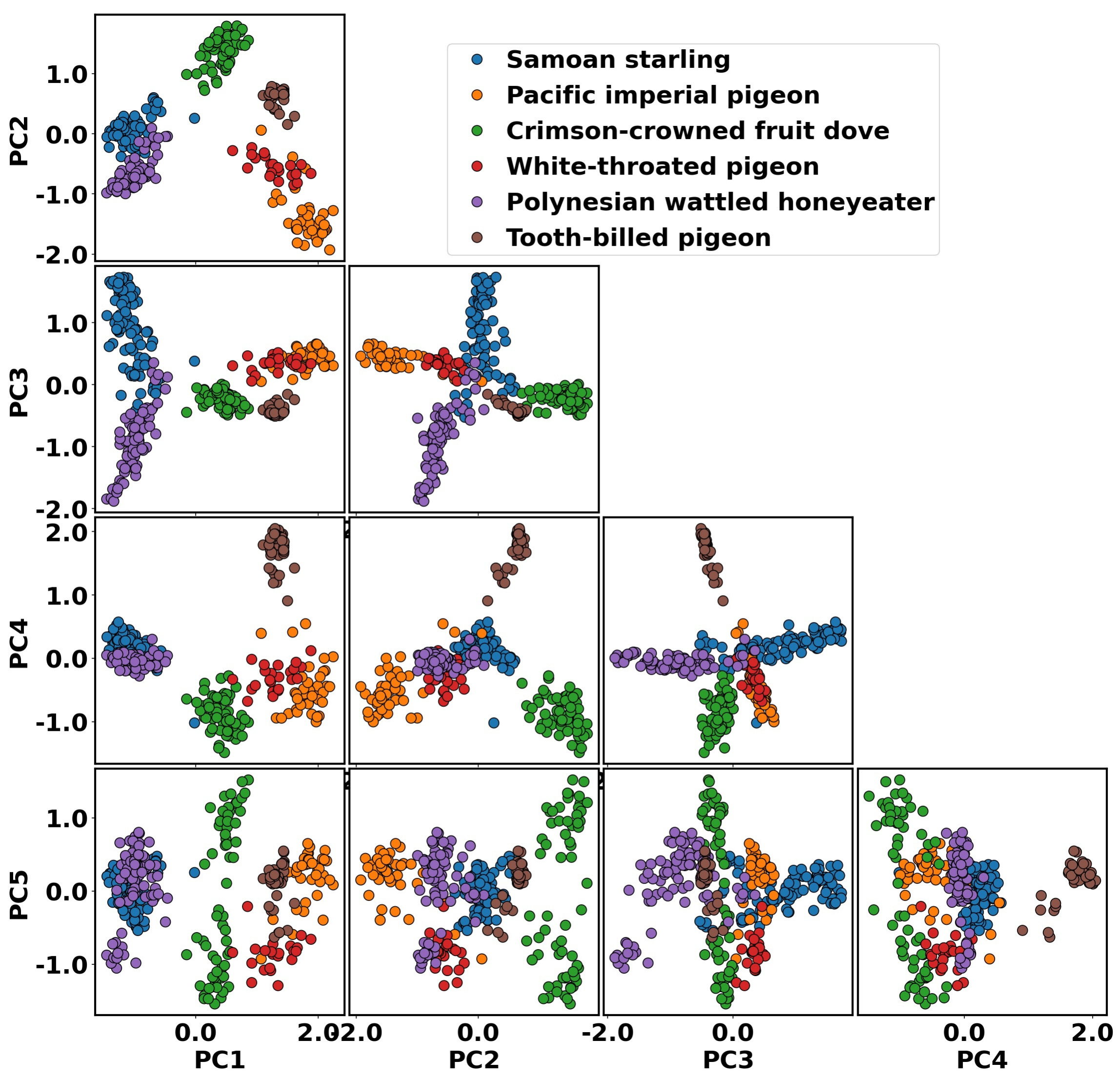}
    \caption{Scatter plot matrix of the top 5 principal components of the Perch model embeddings.}
    \label{fig:perch_cluster}
\end{figure}

Figure \ref{fig:perch_cluster} illustrates the top five PCA components for Perch, providing a visual representation of the clustering patterns. The distinct clusters observed align with the quantitative performance metrics found for the Perch model.

\subsection{Cosine similarity} 
The evaluation metrics in Table \ref{tab:results} summarize the model's performance on both test datasets. For the Eastern Towhee dataset, the model performed well across all metrics. On the Tooth-Billed Pigeon test set, it achieved a recall of 1.00 and an accuracy of 0.951, correctly identifying all tooth-billed pigeon vocalizations. While the F1 score was 0.778, indicating a higher rate of false positives, this trade-off is acceptable given that maximizing recall is the primary objective. Given a recall of 1.0 and an F1 score of 0.778, the implied precision on the Tooth-Billed Pigeon test set is approximately 0.64. Of the 83 total test clips, the model produced approximately four false positive predictions, that is, non-target clips classified as tooth-billed pigeon. While this increases the manual review effort, a conservationist reviewing flagged clips from a multi-hour deployment would still face a tractable workload.

\begin{table*}[ht]
    \centering
    \begin{tabular}{lcc}
    \toprule
    \textbf{Test Metric} & \textbf{Tooth-Billed Pigeon} & \textbf{Eastern Towhee} \\
    \midrule
    Accuracy & 0.951 & 0.988 \\
    Recall & 1.000 & 0.979 \\
    F1 & 0.778 & 0.968 \\
        \bottomrule
    \end{tabular}
    \caption{Results for classification of the species of interest on the test datasets. Results for the chosen metrics are shown for using a recall of 0.9 to determine the classification threshold in training.}
    \label{tab:results}
\end{table*}

\subsection{Event Detection}
\begin{figure}
    \centering
    \includegraphics[width=0.6\textwidth]{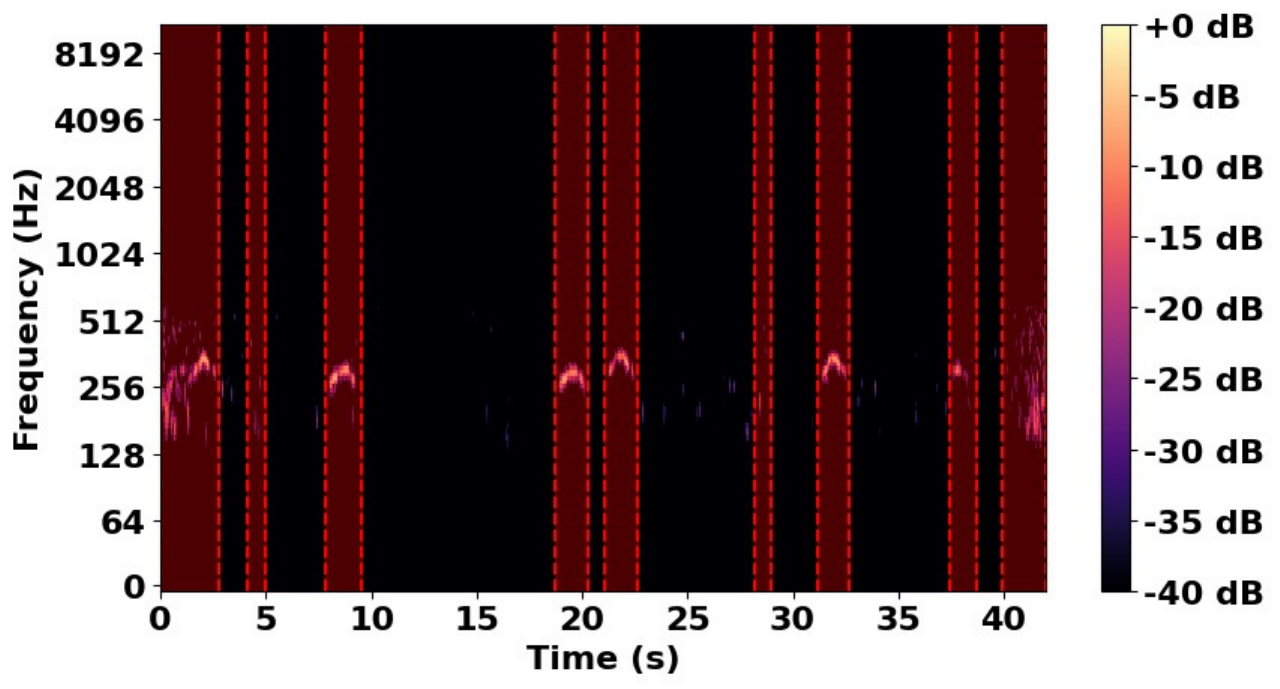}
    \caption{Spectrogram of the test dataset after preprocessing and applying the event detector. The red-highlighted regions indicate the call's detected vocalization events.}
    \label{fig:event_detection}
\end{figure}

The test data is preprocessed as described in Section \ref{sec:preprocessing}. Following this, the data is analyzed with our event detection algorithm. Figure \ref{fig:event_detection} illustrates the results of this detection, with vocalization events highlighted in red. This demonstrates how the event detector is used to isolate calls from a recording.

\section{Discussion} % -----------------------------------------------------
\label{sec:discussions}

This study presents a largely automated pipeline, incorporating targeted manual quality control steps, for classifying rare bird species that have not been previously represented in classification models. The manual quality-control steps, namely spectrogram inspection, overlap removal, and threshold calibration, are confined to the one-time development phase; once calibrated, the detection pipeline runs automatically on new field recordings, with manual effort limited to reviewing flagged candidate detections. The results demonstrate a practical, easy-to-use approach that either outperforms previous one-shot classifiers \citep{Moummad2024, Poutaraud2023} or achieves comparable performance \citep{Ghani2023}. Notably, the pipeline is capable of training an effective classifier from just five minutes of recordings, achieving a recall of 0.98 and an accuracy of 0.90 in a simulated test scenario. When applied to the real-world case of the critically endangered tooth-billed pigeon, the model was trained on only a single recording and achieved a recall of 1.0 and an accuracy of 0.951 on identifying two unseen individuals, highlighting its potential for conservation applications.

% also the entire pipeline is what achieved the performances we're seeing, such as the filtering. 
A key strength of this pipeline is its systematic integration of preprocessing, embedding selection, and classification thresholding, allowing for effective classification even with limited training data. Large-scale bird classifiers, while trained on extensive datasets, often struggle with species that lack sufficient representation. Our approach addresses this challenge by leveraging embeddings from models trained on large datasets while simultaneously optimizing classification for a single target species within a specific acoustic environment. Equally important is the role of preprocessing, including filtering and denoising. To quantify this contribution, we ran an ablation on five tooth-billed pigeon test recordings (Recordings A to E; Table~\ref{tab:detector_ablation}) comprising 32 expert-annotated calls, of which seven were confirmed tooth-billed pigeon calls (three in Recording A and four in the held-out Recording E). The remaining calls were deliberately retained as hard negatives dominated by acoustically similar species (Pacific imperial pigeon, white-throated pigeon, and crimson-crowned fruit dove). For already-segmented calls, bandpass filtering and denoising had little effect on the embedding classifier itself: recall and accuracy were essentially unchanged with or without these steps, reflecting the robustness of the Perch front end. Preprocessing was instead decisive at the event-detection stage. With preprocessing, the wavelet detector produced localized, call-length events (111 in total, median duration on the order of one to two seconds) that cleanly isolated every annotated call. Without preprocessing, the detector collapsed to a single undifferentiated event spanning close to the entire recording in most files (seven events in total, each covering essentially 100\% of its recording), and could no longer localize individual calls. When these large unfiltered events were subsequently windowed into two-second segments and classified, both conditions recovered all seven true tooth-billed pigeon calls with comparable precision on the hard negatives, confirming that the classifier is robust once calls are segmented. Preprocessing is therefore essential to end-to-end performance because it enables reliable event detection and segmentation in field recordings, even though the downstream classifier is itself insensitive to it.

\begin{table}[htbp]
    \centering
    
    \begin{tabular}{lccccc}
        \toprule
        \multirow[b]{2}{*}{\textbf{Recording}} & \multirow[b]{2}{*}{\textbf{Annotated calls}} & \multicolumn{2}{c}{\textbf{With preprocessing}} & \multicolumn{2}{c}{\textbf{Without preprocessing}} \\
        \cmidrule(lr){3-4} \cmidrule(lr){5-6}
         & & Events & Median dur. (s) & Events & Coverage (\%) \\
        \midrule
        A & 3  & 4   & 1.4 & 1 & 97  \\
        B & 10 & 72  & 0.8 & 1 & 100 \\
        C & 9  & 20  & 1.3 & 3 & 98  \\
        D & 6  & 11  & 2.0 & 1 & 100 \\
        E & 4  & 4   & 1.4 & 1 & 99  \\
        \midrule
        Total & 32 & 111 & -- & 7 & -- \\
        \bottomrule
    \end{tabular}
    
    \caption{Event-detector preprocessing ablation on the five tooth-billed pigeon test recordings (A: Salelologa; B: Magiagi; C: Macaulay; D: West Uafato; E: Manumea, held out). With preprocessing (bandpass filtering and denoising) the wavelet detector produces localized, call-length events that isolate every annotated call. Without preprocessing it collapses to one to three events per recording, each covering close to 100\% of the recording, so individual calls can no longer be localized. ``Coverage'' is the fraction of the recording spanned by detected events.}
    \label{tab:detector_ablation}
\end{table}

To compare localization strategies and their computational cost, we evaluated four pipelines on the five test recordings (Table~\ref{tab:localization}): the wavelet event detector and a brute-force sliding-window classifier (2-second windows, 0.5-second hop), each with and without preprocessing. Localization quality was quantified as the mean temporal intersection-over-union (IoU) between detected and annotated call intervals. The event detector with preprocessing localized calls most accurately (mean IoU 0.76, boundary error approximately 0.5 seconds) while issuing only 111 classifier queries across roughly six minutes of audio. Without preprocessing the detector collapses to whole-recording events (IoU 0.04): computationally cheap (7 queries) but unable to localize calls. The sliding-window classifier required 699 queries, a roughly six-fold increase, and three to nine times the runtime, yet localized far less precisely (IoU 0.15 with preprocessing, 0.06 without); on raw audio it additionally missed 22 of the 32 calls and produced 9 false-positive intervals. The event detector combined with preprocessing therefore provides both the most accurate localization and a substantial efficiency gain, reducing the number of embedding computations by roughly six-fold relative to brute-force windowing.

\begin{table}[htbp]
    \centering
    
    \begin{tabular}{lcccc}
        \toprule
        \textbf{Pipeline} & \textbf{Calls recovered} & \textbf{Localization IoU} & \textbf{Classifier queries} & \textbf{Runtime (s)} \\
        \midrule
        Event detector + denoised   & 32/32        & 0.76 & 111 & 84  \\
        Event detector + raw        & 32/32$^{*}$  & 0.04 & 7   & 30  \\
        Sliding window + denoised   & 32/32        & 0.15 & 699 & 286 \\
        Sliding window + raw        & 10/32        & 0.06 & 699 & 270 \\
        \bottomrule
    \end{tabular}
    
    \caption{Localization accuracy and computational cost of four candidate pipelines on the five test recordings (32 annotated calls; $\sim$6 minutes of audio; CPU runtime, one-time model load excluded). Localization IoU is the mean temporal intersection-over-union between detected and annotated call intervals. $^{*}$The raw event detector recovers calls only as a degenerate artifact: it emits single whole-recording events that trivially overlap every call without localizing any of them.}
    \label{tab:localization}
\end{table}

Another finding of this study is that prioritizing recall through a classification threshold, rather than selecting the most confident prediction, significantly improved detection performance. This approach helped mitigate the impact of uncertain or overlapping calls, enhancing the recall of the species of interest even at the cost of increased false positives.

Despite the strong performance of this pipeline, some limitations remain. While the classification threshold is clearly defined, its effectiveness may vary across different species or recording conditions, requiring user adjustment for each new species. In particular, dove and pigeon vocalizations tend to be highly stereotyped, which likely facilitated learning from a single individual in our case study. Species with greater within-individual or between-individual call variability may require additional training examples or wider threshold margins, and the pipeline's generalizability to such species remains to be evaluated. Relatedly, the classification threshold was defined on a single training recording and validated within the same deployment region. Its transferability to recordings from different field sites, recorder hardware configurations, or acoustic environments is not guaranteed, and we recommend recalibrating the threshold when deploying in substantially different conditions. Additionally, although thresholding helps mitigate noise from overlapping calls, source separation techniques could further improve classification by isolating simultaneous vocalizations. Promising methods such as MixIT \citep{Wisdom2020} and other bird call-specific separation algorithms \citep{Dai2021, Xie2024} could be explored in future iterations.

Future work could explore alternative distance metrics, such as Euclidean, Manhattan, or Minkowski, as well as other classification methods like k-nearest neighbors (KNN), decision trees, random forests, neural networks, and autoencoders to further refine performance.

\section{Conclusion} % -----------------------------------------------------
\label{sec:conclusion}

This study presents a novel, largely automated pipeline that enhances few-shot bird call classification, significantly improving the accessibility of rare bird call detection with minimal training data, requiring as little as a single recording. By systematically incorporating preprocessing, embedding selection, and optimized classification thresholds, the pipeline successfully classifies the Tooth-Billed Pigeon, achieving a recall of 1.0 and an accuracy of 0.951, making it suitable for use in the field. Coupling preprocessing with event detection also makes the pipeline efficient: it localizes calls accurately while reducing the number of embedding computations roughly six-fold relative to brute-force windowing, an important consideration when scanning thousands of hours of field audio.
Future work would likely benefit from the integration of source separation techniques and more refined embedding space classification approaches to further refine accuracy and improve precision. More broadly, this pipeline provides a practical and adaptable tool for conservationists monitoring critically endangered species, with potential applications extending to similar conservation challenges.

\section*{Funding} % -----------------------------------------------------

The funding for developing the classification pipeline is from Colossal Biosciences, while the funding for expeditions and data collection is from Toledo Zoo, Birdlife International, and Colossal Biosciences, where the Birdlife International funds are from Waddesdon Foundation via the Zoological Society of London.

\section*{Data Accessibility Statement} %------------------------------------

The raw and processed data supporting the findings of this study are available on Figshare: \url{https://doi.org/10.6084/m9.figshare.29211299.v1}. The code repository (\url{https://github.com/abhishek-jana/few-shot-bird-call}) is archived on Zenodo (\url{https://doi.org/10.5281/zenodo.20726966}) under the MIT license.
The datasets labeled \texttt{us\_train\_data} and \texttt{us\_test\_data} were compiled from the publicly available Xeno-Canto repository. The datasets labeled \texttt{train\_data} and \texttt{test\_data} are a mix of Xeno-Canto samples and field recordings collected during our expeditions. Details regarding the composition and collection of these datasets are described in Sections \ref{sec:eastern_towhee} and \ref{sec:tooth_billed}. Due to conservation concerns and the critically endangered status of the Tooth-Billed Pigeon, we have not made those specific recordings publicly available.

\section*{Competing Interests Statement} %---------------------------------
The authors declare that they have no competing interests.

\section*{Acknowledgments} % ----------------------------------------------- 
We are grateful to the local communities in Samoa who partnered with us to launch the Manumea Friendly Village Program, supporting our efforts to gather essential bird call data. This program aims to deepen our understanding of Manumea vocalizations and improve field detection techniques. Special thanks go to the communities of Magiagi, Faleaseela, Matafaa, Uafato, Malololelei, Tafua-tai, Salelologa, Safotu-Mt Matavanu Crater, Aopo, Falealupo, and Masamasa Falelima National Park, among others, for their invaluable contributions. Additional data were obtained from Xeno-Canto (www.xeno-canto.org), a community-driven repository of bird sound recordings, for which we are also grateful.

We sincerely thank Laulu Fialelei for their expertise and invaluable assistance in the annotation process. Their careful assessment of the field recordings and contributions to call identification were essential in refining the test set and ensuring the accuracy of our classification approach. 

We also thank the CEO and staff of the Samoa Ministry of Natural Resources and Environment, as well as the President and technical teams at the Samoa Conservation Society, Professor Liba Pejchar and Sara Bombaci, and Michael Pardo of the Department of Fish, Wildlife, and Conservation Biology at Colorado State University for their guidance. We further acknowledge the Birdlife International Pacific Partnership in Suva, Fiji, the IUCN Pigeons and Doves Specialist Group, Toledo Zoo.

Additional appreciation goes to our partners who have supported our field data collection over the years, including Auckland Zoo, the New Zealand Department of Conservation, the New Zealand Foreign Affairs and Trade, Pelgar International, the Global Diversity Foundation–Community and Conservation Foundation, and the UK Government’s Darwin Initiative.

We would also like to acknowledge the invaluable knowledge contributed by Dr Ulf Beichle, Dr Gianluca Serra, and Dr Rebecca Stirnemann through their extensive research collaborations on the Manumea.

\section*{Author Contributions} % ----------------------------------------------- 
Abhishek Jana led the development of the classification pipeline, performed data preprocessing, model evaluation, and co-wrote the majority of the manuscript.\\
Moeumu Uili coordinated field data collection, led the expert annotation process, and contributed significantly to the species-specific ecological context and manuscript refinement.\\
James Atherton and Mark O’Brien contributed to conservation strategy design, data interpretation, and manuscript review.\\
Joe Wood was heavily involved in orchestrating the Manumea search, authored key project proposals, and contributed to conservation strategy design and project coordination.\\
Leandra Brickson co-wrote the majority of the manuscript, contributed to project conception, secured funding, supervised the research, and provided critical feedback throughout.\\
All authors reviewed and approved the final manuscript.

\bibliographystyle{Frontiers-Harvard} %  Many Frontiers journals use the Harvard referencing system (Author-date), to find the style and resources for the journal you are submitting to: https://zendesk.frontiersin.org/hc/en-us/articles/360017860337-Frontiers-Reference-Styles-by-Journal. For Humanities and Social Sciences articles please include page numbers in the in-text citations 
\bibliography{references}

\end{document}